# Is Sara Omar the true author of *Dødevaskeren* (*The Dead Washer*)?

## Authorship attribution in Danish literature

Jacques Savoy
University of Neuchatel
2000 Neuchatel, Switzerland
Jacques.Savoy@unine.ch

**Abstract**

The bestseller *Dødevaskeren* (*The Dead Washer*) was published in 2017 under the name Sara Omar. A Danish colleague has suggested that the novel may have been written by another author, possibly Mads Peder Nordbo. To investigate this authorship question, four authorship attribution methods were applied, namely Delta, intertextual distance, Nearest Shrunken Centroids (NSC), and naïve Bayes. Using various parameter settings, these approaches consistently identify Nordbo's writing style in the doubtful novel, although the results do not provide conclusive evidence that he authored the novel in its entirety. In a second step, the novel was divided into non-overlapping segments of 5,000 tokens. Once again, Nordbo's style was identified in the majority of the segments. However, a small number of passages do not clearly support this conclusion and instead suggest the possible involvement of another, as yet unidentified, author, possibly Sara Omar herself.

## 1 Introduction

When a new novel or play is published, we generally assume that the name appearing on the cover corresponds to the actual author. However, there are cases in which the writing style appears to resemble that of another well-known writer. Such authorship attribution enigmas have emerged across different literary traditions, including French literature with the Gary–Ajar question (*La Vie devant soi*, 1975) (Labbé, 2008), Italian literature with the Ferrante question (*L'amica geniale*, 2011) (Tuzzi & Cortelazzo, 2018a, 2018b; Savoy, 2018), and English literature with the Rowling–Galbraith puzzle (*The Cuckoo's Calling*, 2013) (Juola, 2016), not to mention the well-known Shakespeare authorship question (Craig & Kinney, 2009).

In the context of Danish literature, a Danish colleague has suggested that *Dødevaskeren* (*The Dead Washer*), a bestseller published in 2017 under the name of Sara Omar, may have been written, at least partially, by another writer. The novel achieved considerable success and has been translated into Swedish, Norwegian, Serbian, Macedonian, and French. Sara Omar is a Danish-Kurdish writer, human rights activist, and poet. Thus, she is not an unknown figure; however, she has published primarily in other genres, notably poetry and newspaper articles.

Because the term “author” can refer to different situations, its meaning should first be clarified. In the remainder of this manuscript, the term refers to the person who actually writes the text, rather than to someone who provides the blueprint, contributes some dialogue, or establishes the main psychological traits of certain characters.

This authorship question can be examined from different perspectives. The first approach is to assume that the novel (hereafter referred to as Sara’s novel) is the work of a single person, namely Sara Omar. However, we do not have a novel written by Sara Omar in the same genre as the doubtful text. Moreover, poems, plays (whether in verse or prose), and novels correspond to different genres, each of which can have a substantial influence on writing style (Savoy, 2020).

As a second line of inquiry, we could assume that this bestseller is the work of a single author, but not Sara Omar. To facilitate this investigation, we have a list of 15 Danish novelists and, for each of them, two novels belonging to either the crime or contemporary fiction subgenre. Under this assumption, different authorship attribution models can be applied to determine whether the stylistic fingerprint of one of these Danish authors can be detected in Sara’s novel.

Third, *Dødevaskeren* could have been written by two distinct authors. In this case, we assume that one of the authors is Sara Omar and that the second is a writer represented in our Danish corpus. To evaluate this hypothesis, traditional authorship attribution models must be adapted by considering different passages of the doubtful text separately.

The remainder of this paper is organized as follows. The next section describes the Danish corpus used in our investigations. Section 3 presents the results obtained using the Delta model, our first authorship attribution approach. Labbé’s intertextual distance, our second authorship identification approach, is described in Section 4. Section 5 describes and evaluates two additional authorship attribution methods based on machine learning techniques, namely Nearest Shrunken Centroids (NSC) and naïve Bayes. Section 6 applies rolling Delta to investigate whether the doubtful novel was written by one or two authors. Some linguistic considerations that may provide additional support for our findings are presented in Section 7. Finally, the conclusion summarizes the main findings of this study.

## 2 Corpus of Danish Novels

To provide a solid basis for this study, a Danish corpus was compiled by Danish colleague. Table 1 presents the 15 Danish writers included in the corpus, together with the length of their novels. The values reported in the last three columns of Table 1 correspond to the concatenation of all the books written by each author. In the column labelled “Length (in tokens),” the value is computed by considering as tokens all words[1] (i.e., sequences of letters), punctuation symbols, numbers, and terms composed of both letters and numbers (e.g., IBM360). In the next column, only words, defined as sequences of letters, are counted, regardless of whether they appear in uppercase or lowercase. The final column reports the vocabulary size, that is, the total number of distinct word-types used by each author.

[1] Because the term “word” is ambiguous, some clarification is necessary. In the sentence “I saw a man with a saw,” there are seven word-tokens (or simply tokens), but only five distinct word-types (or types), namely {I, saw, a, man, with}.

The last two rows provide information about two specific novels. The penultimate row corresponds to a novel known to have been written by two authors, Blædel and Nordbo, whereas the final row corresponds to the doubtful text. More detailed information about each novel is provided in Table A.1 in the Appendix.

**Table 1.** Statistics about the novelists occurring in the corpus

| Author name | First name | Genre | Length (in tokens) | Length (in words) | Vocabulary |
|---|---|---|---|---|---|
| Blædel | Sara | F | 213,326 | 181,738 | 181738 |
| Conradsen | Keld | M | 244,840 | 210,427 | 16,598 |
| Gazan | Sissel-Jo | F | 214,291 | 174,354 | 13,841 |
| Gerhardt | Maria | F | 56,627 | 47,011 | 7,567 |
| Guldager | Katrine | F | 194,482 | 164,053 | 14,301 |
| Halfdan | Finn | M | 281,606 | 226,622 | 16,769 |
| Harvig | Steen | M | 186,570 | 151,245 | 13,128 |
| Høeg | Tine | F | 40,779 | 38,751 | 5,973 |
| Jensen | Martin | M | 219,417 | 180,652 | 13,818 |
| Jessen | Ida | F | 87,016 | 72,563 | 8,331 |
| Korsgaard | Thomas | M | 165,634 | 131,043 | 9,403 |
| Nordbo | Mads Peder | M | 325,130 | 261,289 | 12,810 |
| Petersen | Eddie Thomas | M | 183,674 | 158,065 | 14,393 |
| Sabroe | Morten | M | 222,426 | 177,688 | 13,376 |
| Tellerup | Malte | M | 99,079 | 81,358 | 9,712 |
| Blædel-Nordbo | Sara / Mads Peder | F / M | 110,110 | 89,135 | 8,097 |
| Omar | Sara | F | 100,300 | 79,660 | 6,926 |

Several important considerations should be kept in mind. First, the corpus contains approximately 2,945,307 tokens, a quantity roughly ten times greater than the length of the longest English novel, Melville's *Moby Dick*. Second, all the novels were published within a relatively limited time span of approximately 10 years. Third, the text genres represented in the corpus are broadly comparable, as the books consist mainly of novels intended for adult readers (some subgenres per book are given in Table A.1). Fourth, each novel contains more than 10,000 tokens, a sufficiently large number to achieve a relatively high degree of reliability with the selected authorship attribution models.

Before analyzing this corpus, some preprocessing is required. First, all uppercase letters are converted to lowercase. For example, *Og* is converted to *og* (and). Our tokenizer is able to identify words written entirely in uppercase letters (e.g., USA, WHO, DENMARK), which are left unchanged. The same procedure is applied to word-types that consistently occur in capitalized form, such as Matthew, Marie, or Elliot. As these examples illustrate, such word-types are usually proper names.

As Danish is not as frequently employed in natural language processing as some other languages, some general linguistic information may be useful. Danish belongs to the North Germanic branch of the Germanic language family, together with Norwegian and Swedish. It has a relatively rich inflectional morphology (e.g., *bil/biler* (car/cars)). Its basic sentence structure follows the SVO syntactic pattern, with the verb usually occurring in second position. Danish has two grammatical genders, common and neuter, and two numbers, singular and plural. The plural is often marked by the suffixes *-er* or *-e*. A distinctive feature of Danish is that definite determiner can be marked by a suffix attached to the noun, in both the singular and the plural, as illustrated by *bilen* (the car) and *bilerne* (the cars).

As in all natural languages, exceptions occur, for example, in noun inflection (*fod/fødder* (foot/feet)) and verb conjugation (*tager/tog* (takes/took)). Lexical ambiguities can also arise. For instance, *I* can function either as a pronoun (the second-person plural form) or as a preposition[2]. Similarly, *det* can function either as a determiner or as a pronoun. Three additional characters are particularly characteristic of Danish orthography: *ø* (historically related to *oe*), *å* (historically related to *aa*), and *æ*.

Some overall lexical statistics provide an overview of the lexical characteristics represented in this corpus. The 20 most frequent word-types in Danish are {*og, det, i, at, han, på, jeg, var, en, er, ikke, til, hun, der, med, så, den, af, for, havde*}. As can be observed, these items generally do not convey specific lexical or topical meaning. Rather, they belong to functional categories, including determiners (*det, en*), prepositions (*af, i*), conjunctions (*og, eller*), pronouns (*jeg, dig, han, hun*[3]), and auxiliary or highly frequent verb forms (*var*, *havde*). The linguistic fingerprint of an author is often strongly associated with such closed-class words[4], because their use is largely unconscious and is less directly related to the topic or content of a document.

Finally, to obtain an initial overview of the stylistic differences among the Danish novelists, Figure 1 presents a Principal Component Analysis (PCA) projection based on the 200 most frequent words (MFWs) (Binongo & Smith, 1999), (Craig & Kinney, 2009), (Tuzzi & Cortelazzo, 2018a). Figure 1 is based on the first two principal components, which account for 48.2% and 11.9% of the total variation among the novelists, respectively. Thus, the two-dimensional representation captures 60.1% of the overall variation, while the remaining 39.9% is not represented in the figure.

It is important to note, however, that this representation is not orthonormal, meaning that the units of distance are not equivalent along the two axes. In the present case, the horizontal axis spans a larger range of variation than the vertical axis. Consequently, visual distances between points should be interpreted with some caution.

---

[2] The tokenizer detects *I* (pronoun) and *i* (preposition) and thus transforms all *I* (pronoun) into lowercase letter (as the pattern 'The' and 'the' that occurs in English).

[3] As already observed in other corpora, the masculine form occurs more frequently than the feminine form.

[4] The term "closed class" refers to grammatical categories for which the creation of new words is generally not possible.

In this figure, the horizontal axis primarily indicates a contrast between the frequent use of first-person personal pronouns (*jeg*, *mig*, *vi*) on the left and third-person pronouns (*han*, *hun*) on the right. The vertical axis indicates that, in the upward direction, authors tend to use, among other terms, the past-tense forms *var* (“was”), *havde* (“had”), and *sagde* (“said”) more frequently. In the opposite direction, writers tend to overuse the present-tense forms of these verbs, such as *er* (“is”) and *siger* (“says”).

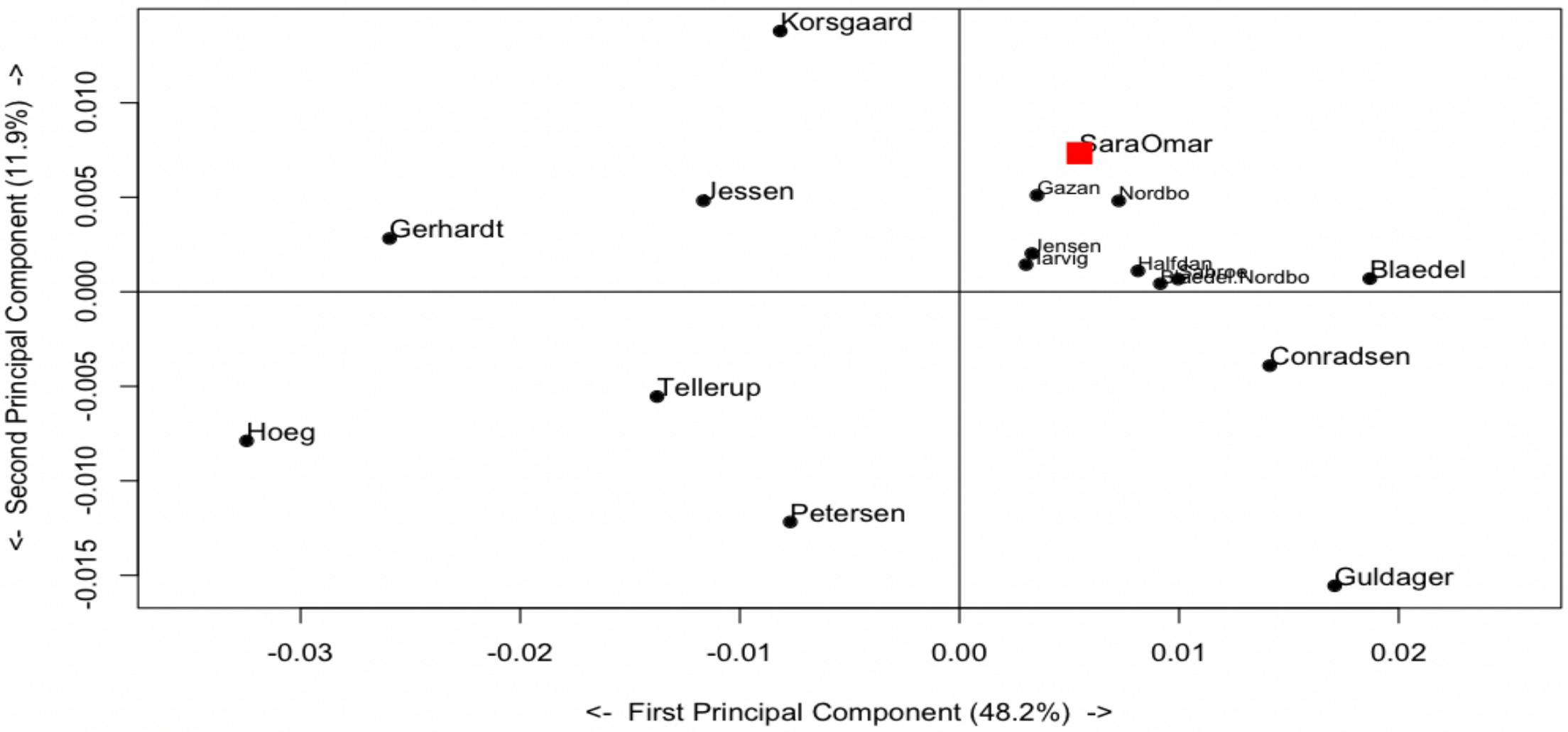

**Figure 1.** PCA projection of Danish novelists based on 200 MFWs

Several authors appear stylistically close to Sara’s novel, represented by the square. Moreover, this point is relatively close to the center of the projection, which can be interpreted as representing the average position of the Danish author profiles in this analysis. Based on the Euclidean distances calculated in the original PCA space, Gazan is the closest author (distance = 0.0030), followed by Nordbo (0.0031) and Jensen (0.0057). In contrast, several authors—including Høeg, Gerhardt, Guldager, Petersen, Tellrup, and Jessen—exhibit substantially greater distances from Sara’s novel.

Overall, the PCA provides a useful preliminary visualization of stylistic variation within the corpus. However, these results should be regarded as descriptive rather than as direct evidence of authorship, since the first two principal components account for only part of the total variation.

## 3 Authorship Attribution Based on Delta

To determine the authorship of a text, numerous authorship attribution approaches have been proposed. For example, Juola and Vescovi (2011) identify more than 1,000 different models and parameter settings. It is therefore difficult to assume that a single attribution model can consistently provide the correct answer under all circumstances. According to the *no free lunch theorem* (Wolpert, 2001), no learning algorithm can be universally superior to all others. Consequently, in order to establish a proposition with a higher degree of confidence, several models, preferably with different parameter settings, should be considered.

To meet these requirements, the well-known Delta authorship attribution model has been selected. Originally proposed by Burrows (2002), several variants have subsequently been proposed and evaluated (Hoover, 2004a, 2004b; Burrows, 2007; Hoover, 2007), while its theoretical properties have also been investigated (Argamon, 2008). The model has been employed in numerous authorship attribution studies (Jockers *et al.*, 2008; Savoy, 2012, 2016; Eder, 2015; Evert *et al.*, 2017; Kocher & Savoy, 2018). Moreover, Jockers and Witten (2010) showed that the Delta method can outperform Support Vector Machine (SVM) approaches in authorship attribution.

This attribution model makes use of a subset of the vocabulary and considers only the most frequent word-types (MFWs), which consist mainly of function words such as determiners, pronouns, prepositions, conjunctions, and some auxiliary verb forms. The number of terms to be included (e.g., word-types or short sequences of letters) is not precisely defined; however, values between 200 and 500 terms are commonly used. These terms are selected without making any explicit reference to the doubtful text.

For each selected term $t_i$, its relative term frequency, denoted $rtf_{ij}$, is computed for each text $T_j$. This value is then standardized using the mean and standard deviation of the corresponding term across all texts in the reference corpus. The resulting standardized value, denoted Z score$(t_{ij})$, is defined according to Equation 1.

$$\text{Z score}\left(t_{ij}\right) = \left(\text{rtf}_{ij} - \overline{rtf_i}\right) / s_i \qquad (1)$$

Then, given a query text $Q$, an author profile $A_k$ (defined as the concatenation of all writings by author $k$), and a set of terms $t_i$, for $i = 1, 2, \ldots, m$, the Delta distance between $Q$ and $A_k$, denoted $D(Q, A_k)$, is computed according to Equation 2.

$$\Delta(Q, A_k) = 1/m \cdot \sum_{i=1}^{m} \left| \, Z\,score\left(t_{iQ}\right) - Zscore\left(t_{ik}\right) \right| \qquad (2)$$

When, for a given term, the two $Z$ scores are large and have opposite signs, the distance increases. In this situation, one author tends to use the corresponding word more frequently than the corpus mean, whereas the other tends to use it less frequently. Conversely, when the $Z$ score values are very similar for all terms, the distance between the two texts is reduced. The author profile $A_k$ yielding the smallest distance, among all writers is identified as the most likely author of the query text.

To apply the Delta model to the present corpus, the stylistic feature set must first be defined. More precisely, the $m$ most frequent word-types (MFWs) are selected, with $m$ ranging from 200 to 800. Previous empirical studies suggest that satisfactory results can generally be obtained with values of $m$ equal to or greater than 200. To define the MFWs, all books included in the corpus are taken into consideration.

The Delta model has no prior knowledge of the possible author of the query text. Consequently, it ranks all items (e.g., books, author profiles, etc.) included in the input set according to their distance from the query text. In the results presented below, only the five highest-ranked candidates are discussed.

In this approach, the Manhattan distance measure has been selected because of its good empirical performance in previous studies. The distance can take values ranging from 0 (e.g., the distance between a text and itself) to positive values that could be greater than 1.0. When more stylistic features are considered, the resulting Manhattan distance will increase simply because more dimensions are taken into account (see Equation 2). Consequently, the absolute value of the distance cannot be interpreted directly. For example, a distance of 0.65 may be considered either large or small, depending on the other distance values obtained. The appropriate interpretation is therefore to compare a distance value with the other values within the same ranking.

To facilitate interpretation, the current implementation of the Delta model provides an *estimate* of the probability that each candidate is the correct author, under the *strong assumption* that the true author is represented in the corpus (Savoy, 2016).

Table 2a reports the five highest-ranked candidates produced by the Delta model using 200 MFWs, with Sara's novel as the query document. In this instance, each novel corresponds to a separate entry in the input list, thus implementing an instance-based approach. The second column reports the Delta distance from Sara's novel, while the third column indicates the percentage change relative to the distance at the preceding rank. The next column provides the identifier of the corresponding novel.

**Table 2a.** Ranked lists achieved by the Delta (200 MFWs, instance-based) to Sara's novel

| Rank | Distance | % Change | Novel | Prob. |
|---|---|---|---|---|
| 1 | 0.622 | | Nordbo3 | 0.544 |
| 2 | 0.674 | 8.4% | Nordbo2 | 0.192 |
| 3 | 0.694 | 2.9% | Nordbo1 | 0.131 |
| 4 | 0.696 | 0.4% | Blædel-Nordbo | 0.124 |
| 5 | 0.899 | 29.1% | Halfdan2 | 0.002 |

**Table 2b.** Ranked lists achieved by the Delta (500 MFWs, instance-based) to Sara's novel

| Rank | Distance | % Change | Novel | Prob. |
|---|---|---|---|---|
| 1 | 0.748 | | Nordbo3 | 0.347 |
| 2 | 0.763 | 1.9% | Nordbo1 | 0.260 |
| 3 | 0.767 | 0.6% | Nordbo2 | 0.237 |
| 4 | 0.798 | 4.0% | Blædel-Nordbo | 0.128 |
| 5 | 0.942 | 18.0% | Sabroe1 | 0.007 |

Overall, the three novels by Nordbo occupy the top positions in the ranking, indicating a strong stylistic similarity to the doubtful novel. Moreover, the book co-authored by Nordbo also appears high in the ranking. This finding may suggest that Nordbo was responsible for a substantial part of the writing process in this co-authored novel. Alternatively, it may indicate that Blædel's style is particularly close to Nordbo's, although Figure 1 provides little support for this possibility. Further evidence can be obtained from the large increase in the percentage change between the fourth and fifth ranks in Table 2a. Once Nordbo's style is no longer represented among the highest-ranked candidates, the Delta distance increases markedly.

Table 2b presents the same information, but using 500 MFWs. Considering an additional parameter setting provides a further assessment of the robustness of the attribution results. As shown in the table, the ranking differs somewhat from that obtained with 200 MFWs, but the main findings remain unchanged. The three novels by Nordbo occupy the highest positions, followed by the novel he co-authored. These results therefore provide further support for Nordbo's involvement. As in the previous table, the percentage change in the Delta distance increases substantially between the fourth and fifth positions, providing additional support for the hypothesis that Nordbo may have contributed to the writing of Sara's novel. Similar results are obtained when using 250, 300, or 800 MFWs.

Overall, Tables 2a and 2b yield similar results. The novels most stylistically similar to Sara's novel are those authored or co-authored by Nordbo. As soon as the Delta model considers a novel written by another author, the distance increases and the corresponding probability decreases substantially. Although the estimated probabilities do not always provide a clear indication for an individual novel, the attribution pattern becomes more pronounced when all novels by a given author are considered together.

Following this perspective, Tables 3a and 3b present the rankings obtained by considering the authors directly, using a profile-based approach. The Delta model then ranks the authors according to their stylistic similarity to Sara's novel. The results are consistent with those reported above. Among the 15 novelists represented in our corpus, Nordbo has the profile closest to Sara's novel. Moreover, the percentage changes increase substantially when an author other than Nordbo is considered (e.g., +42.9% in Table 3a). The estimated probabilities likewise consistently identify Nordbo as the most likely author.

**Table 3a.** Ranked lists achieved by the Delta (200 MFWs, profile-based) to Sara's novel

| Rank | Distance | % Change | Novel | Prob. |
|---|---|---|---|---|
| 1 | 0.691 | | Nordbo | 0.990 |
| 2 | 0.987 | 42.9% | Jensen | 0.003 |
| 3 | 0.988 | 0.1% | Halfdan | 0.003 |
| 4 | 1.013 | 2.4% | Sabroe | 0.002 |
| 5 | 1.024 | 1.1% | Korsgaard | 0.001 |

A few important remarks are necessary at this point. The conclusion that "Nordbo is the true author behind Sara's novel" cannot yet be drawn. First, the Delta model bases its attribution on a corpus containing only 15 novelists. The actual author could therefore be an unknown writer who is not represented in the corpus. A second possibility is that Nordbo contributed to the writing process but was responsible for only a portion of Sara's novel.

To evaluate the hypothesis that Nordbo is the sole or principal author of the doubtful novel, a verification procedure proposed by Savoy (2018) can be applied. The proposed author, Nordbo, is removed from the evaluation. If Nordbo indeed contributed substantially to the writing process, the distance values obtained by the Delta model should be relatively similar across the remaining authors, with no single candidate emerging as clearly more likely than the others. In other words, the percentage changes between adjacent ranks should remain relatively small.

**Table 3b.** Ranked lists achieved by the Delta (500 MFWs, profile-based) to Sara's novel

| Rank | Distance | % Change | Novel | Prob. |
|---|---|---|---|---|
| 1 | 0.815 | | Nordbo | 0.975 |
| 2 | 1.062 | 30.2% | Sabroe | 0.007 |
| 3 | 1.076 | 1.4% | Korsgaard | 0.005 |
| 4 | 1.078 | 0.2% | Jensen | 0.005 |
| 5 | 1.108 | 2.7% | Halfdan | 0.003 |

The results of this experiment are reported in Tables 4a and 4b, using the profile-based approach with 200 and 500 MFWs, respectively. These tables show that, after Nordbo is removed, the five highest-ranked authors have relatively similar stylistic distances from Sara's novel. None of them can be clearly identified as a plausible author of the doubtful text. The percentage changes between successive ranks also remain consistently small, indicating that none of these authors exhibits a particularly strong stylistic similarity to Sara's novel. Furthermore, the rankings differ slightly between Tables 4a and 4b, despite the fact that the two evaluations use otherwise identical conditions and differ only in the number of MFWs considered (200 versus 500). This variation provides further support for the hypothesis that, once Nordbo is excluded, none of the authors represented in the corpus emerges as a strong candidate for authorship of the doubtful novel.

**Table 4a.** Ranked lists achieved by the Delta (200 MFWs, profile-based) to Sara's novel (without Nordbo)

| Rank | Distance | % Change | Novel | Prob. |
|---|---|---|---|---|
| 1 | 1.025 | | Halfdan | 0.375 |
| 2 | 1.052 | 2.7% | Sabroe | 0.215 |
| 3 | 1.085 | 3.0% | Jessen | 0.113 |
| 4 | 1.087 | 0.2% | Korsgaard | 0.108 |
| 5 | 1.121 | 3.1% | Gazan | 0.055 |

**Table 4b.** Ranked lists achieved by the Delta (500 MFWs, profile-based) to Sara's novel (without Nordbo)

| Rank | Distance | % Change | Novel | Prob. |
|---|---|---|---|---|
| 1 | 1.104 | | Sabroe | 0.405 |
| 2 | 1.146 | 3.8% | Halfdan | 0.174 |
| 3 | 1.159 | 1.1% | Korsgaard | 0.135 |
| 4 | 1.163 | 0.4% | Jessen | 0.124 |
| 5 | 1.215 | 4.5% | Jensen | 0.044 |

## 4 Authorship Attribution Based on Intertextual Distance

As a second authorship attribution model, the intertextual distance measure proposed by Labbé (2007) was selected. The effectiveness of this model has been evaluated in several studies covering different contexts (Labbé, 2008; Tuzzi, 2010; Savoy, 2012; Kocher & Savoy, 2017; Tuzzi

& Cortelazzo, 2018). For example, Labbé and Labbé (2013) applied this distance measure to detect duplicate and automatically generated scientific articles.

The intertextual distance returns a value between 0.0 and 1.0, depending on the lexical overlap between two documents. When two texts are identical, the distance is 0.0. Conversely, the maximum distance of 1.0 is obtained when the two texts have no terms in common (e.g., when one is written in Danish and the other in Chinese). Between these two extremes, the distance depends on the number of terms occurring in both texts and on their respective frequencies. For example, a distance of 0.3 indicates that, after taking occurrence frequencies into account, approximately 30% of the lexical content differs between the two texts.

More precisely, following Labbé (2007), the distance between Text A and Text B, denoted *D(A,B)*, is computed according to Equation 3. In this equation, $n_A$ denotes the length of Text A, measured in tokens, and $tf_{iA}$ denotes the absolute frequency of the $i$th term, for $i = 1, 2, \ldots, m$. The value $m$ represents the size of the vocabulary. Because the two texts rarely have the same length, let us assume that Text B is the longer of the two. To make the two texts comparable, Text B is reduced to the length of the shorter text by multiplying each term frequency (in our case, $tf_{iB}$) by the ratio of the two text lengths, as indicated in the second part of Equation 3. As a variant, terms occurring only once (hapax legomena) may be removed before computing the distance.

$$D(A,B) = \sum_{i=1}^{m} \left| tf_{iA} - \widehat{tf_{iB}} \right| \Big/ (2 \cdot n_A) \qquad \text{with } \widehat{tf_{iB}} = tf_{iB} \cdot n_A / n_B \qquad (3)$$

To begin the evaluation, the distances between every pair of novels were computed. The resulting matrix contains 31 rows and 31 columns, corresponding to 961 distance values. Since the distance between A and B is identical to that between B and A, only half of these values are unique. This large number of values makes it difficult to identify the most relevant patterns directly. However, the resulting distances can be represented graphically in the form of a dendrogram, as shown in Figure 2.

The vertical axis on the left represents the intertextual distance between clusters, that is, groups of novels. Moving from bottom to top, the intertextual distance increases, while the stylistic similarity between the corresponding novels or groups of novels decreases. To construct the dendrogram, the system progressively merges individual novels and clusters until all novels are incorporated into a single cluster. Consequently, as the intertextual distance becomes relatively large, the resulting groupings become less informative for the present analysis.

The most relevant information is displayed in the lower part of Figure 2. To facilitate interpretation, a first dotted line is drawn at a distance of 0.2 and a second at 0.3. When the intertextual distance between two texts is below 0.2, the probability that the two documents were written by the same author is estimated at 0.998. In our corpus, this criterion is satisfied for Korsgaard and Nordbo and is approached by Blædel (0.217) and Sabroe (0.230).

For distances between 0.2 and 0.3, several pairs of texts written by the same author can be observed (e.g., Jessen, Conradsen, Gazan, Harvig, and Halfdan). In such cases, authors may intentionally adopt a different perspective or slightly modify the underlying genre of the text, for example, by moving from a historical novel toward an adventure novel.

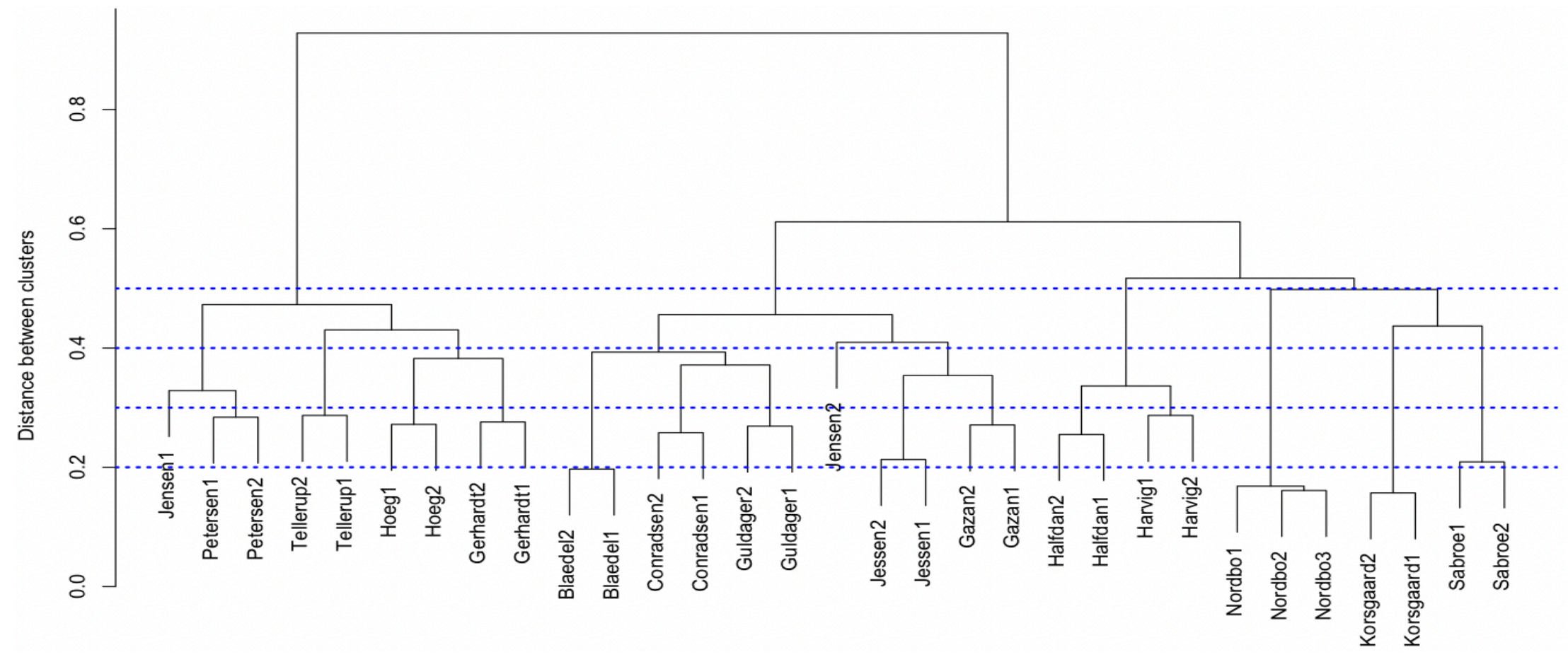

**Figure 2.** Labbé's distance using all the novels without the doubtful one (Ward method)

As the difference in text genre increases, for example, from a crime novel to a romance novel, the resulting intertextual distance also tends to increase. Similarly, when the publication dates of two documents differ by more than 20 years, the intertextual distance may increase even when both texts were written by the same author.

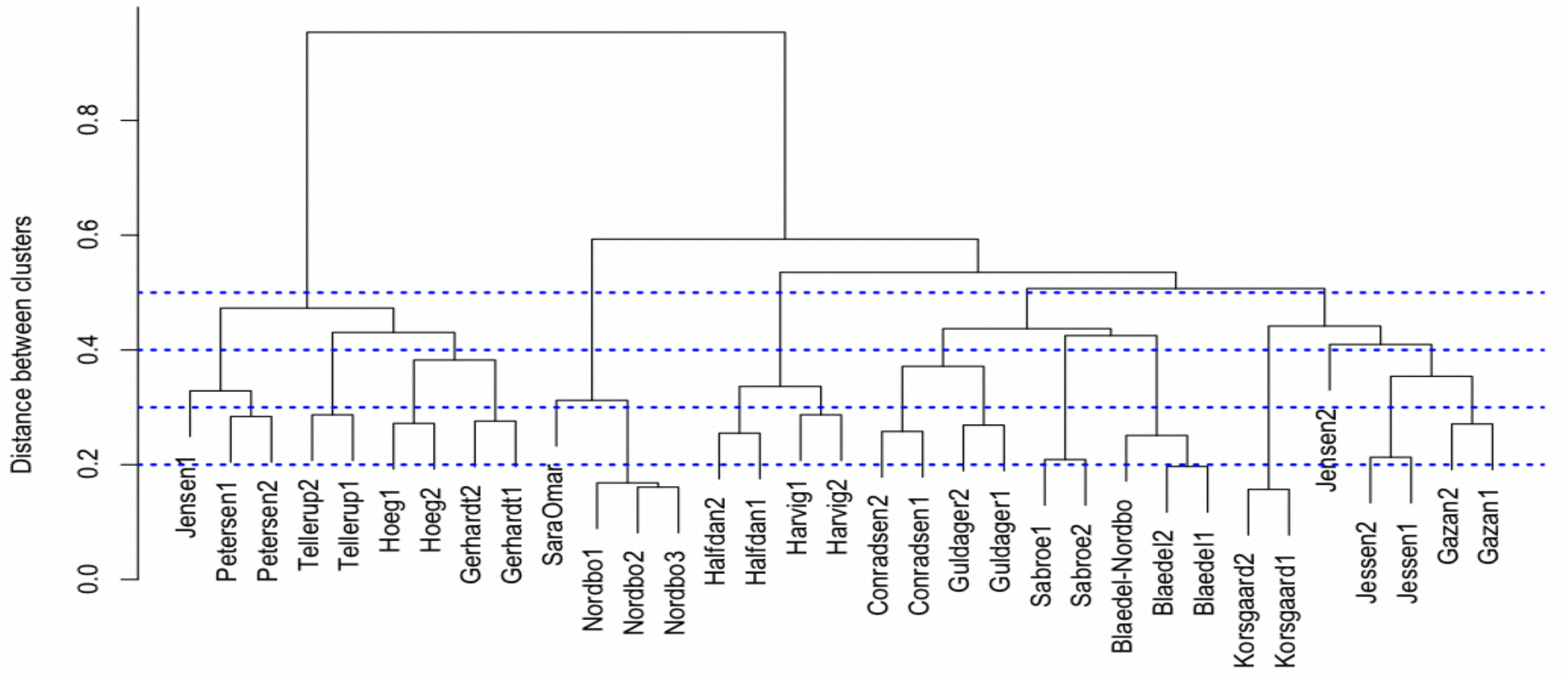

**Figure 3.** Labbé's distance using all the novels (Ward method)

In a second step, the novel co-authored by Blædel and Nordbo and the doubtful novel were incorporated into the analysis, and the resulting dendrogram is shown in Figure 3. The co-authored novel is positioned close to Blædel's group, while the doubtful novel appears close to Nordbo's cluster, although its intertextual distance from this cluster is greater than 0.2. However, novels written by the same author can also exhibit intertextual distances above 0.2, as illustrated, for example, by Tellerup and Høeg. When two texts are written by the same author, a distance between 0.2 and 0.3 may reflect a change in genre or subgenre, for example, from comedy to tragedy, even when both texts are written in prose.

**Table 5.** Ranked list of Labbé's intertextual distances between two novels

| Rank | Distance | Title | Title |
|---|---|---|---|
| 1 | 0.177 | Nordbo2 | Nordbo3 |
| 2 | 0.179 | Nordbo2 | Nordbo1 |
| 3 | 0.181 | Korsgaard2 | Korsgaard1 |
| 4 | 0.187 | Nordbo3 | Nordbo1 |
| 5 | 0.217 | Blædel2 | Blædel1 |
| 6 | 0.230 | Sabroe1 | Sabroe2 |
| 7 | 0.254 | Blædel2 | Blædel-Nordbo |
| 8 | 0.257 | Jessen2 | Jessen1 |
| 9 | 0.258 | Blædel1 | Blædel-Nordbo |
| 10 | 0.258 | Nordbo3 | Blædel-Nordbo |
| 11 | 0.262 | Nordbo1 | Blædel-Nordbo |
| 12 | 0.269 | Nordbo2 | Blædel-Nordbo |
| 13 | 0.273 | Halfdan2 | Halfdan1 |
| 14 | 0.276 | *Nordbo1* | *SaraOmar* |
| 15 | 0.278 | *Nordbo3* | *SaraOmar* |
| 16 | 0.279 | Conradsen2 | Conradsen1 |
| 17 | 0.282 | *Nordbo2* | *SaraOmar* |
| 18 | 0.283 | Gazan2 | Gazan1 |
| 19 | 0.289 | *Blædel-Nordbo* | *SaraOmar* |
| 20 | 0.290 | Gerhardt2 | Gerhardt1 |
| … | … | … | … |

A closer examination of the intertextual distances between pairs of novels can provide additional relevant information. Table 5 reports the 20 smallest distances. As can be observed, all pairs of books with a distance less than or equal to 0.2 were written by the same author. As the distance increases, pairs involving Nordbo and the Blædel–Nordbo co-authored novel begin to appear. The resulting distance values are not particularly large. If Nordbo were the sole author of the Blædel–Nordbo novel, these distances, ranging from 0.258 to 0.269, would be consistent with values observed between novels belonging to the same genre but written by different authors.

A second interpretation is also possible. Nordbo's novels and the Blædel–Nordbo co-authored novel belong to the same genre (e.g., crime fiction), and the writing process may have involved close collaboration between the two authors. This could explain the relatively small intertextual distances observed between Nordbo's novels and the co-authored book (see Table 5).

At slightly larger intertextual distances, pairs involving Nordbo and Sara's novel can also be observed, with values ranging from 0.276 to 0.282. However, this raises the same question encountered previously with the Blædel–Nordbo novel: does Sara's novel belong to the same subgenre as Nordbo's novels? The available evidence suggests that it does not. If Sara's novel belongs to a different genre or subgenre from Nordbo's three novels, this difference could account for the somewhat larger distances and may nevertheless be consistent with Nordbo's involvement in the writing process, although not necessarily as the sole author.

Alternatively, Nordbo could have been the sole author of the doubtful novel. Under this hypothesis, one would expect a clear shift in topic, genre, or subgenre between Nordbo's three novels represented in the corpus and the doubtful novel. Such a shift could account for the somewhat larger intertextual distances while remaining compatible with common authorship.

Thus, the intertextual distance results provide evidence of a stylistic relationship between Nordbo and Sara's novel, but they do not, on their own, allow us to determine whether Nordbo was the sole author or whether the novel resulted from a collaboration between Nordbo and Sara.

## 5 Machine Learning Approaches

The third method considered is the Nearest Shrunken Centroids (NSC) method (Tibshirani *et al.*, 2002, 2003), which has been shown to be effective for authorship attribution (Jockers, 2013; Jockers & Witten, 2010; Kocher & Savoy, 2018). This approach can be viewed as a variant of nearest-centroid classification in which less discriminative features are downweighed or effectively ignored. Specifically, small feature weights are shrunk toward zero, an idea related to regularization techniques such as ridge and lasso regression. Without providing all the methodological details (see Tibshirani *et al.*, 2003), the general procedure is as follows.

For each term $t_i$ (where $i$ =1, 2, …, $m$) appearing in a novel $T_j$ = 1, 2, … $s$, its relative term frequency $rtf_{ij}$ is estimated. From these values, the mean relative term frequency across all novels in the corpus, denoted by (denoted $\overline{rtf_i}$) is computed, as well as the mean relative term frequency across all novels written by author $A_k$ (for $k$ = 1, 2, … $r$) (indicated by $\overline{rtf_{ik}}$). Let $n_k$ denote the number of novels written by author $A_k$. The standard deviation $s_i$ is then determined. These quantities are used to calculate a discriminative value $w_{ik}$ for term $t_i$ and author $A_k$, according to Equation 4.

$$\mathrm{w}_{ik} = \left(\overline{\mathrm{rtf}_{ik}} - \overline{rtf_i}\right) / (m_k \cdot s_i) \qquad \text{with } m_k = \sqrt{1/n_k + 1/n} \tag{4}$$

The principle underlying this feature-weighting scheme can be explained as follows. When the mean frequency of the $i$th term for the $k$th author is similar to the overall mean frequency of this term, the resulting feature weight for this author is small. This indicates that the term has little discriminative power for distinguishing this author from the other authors. However, the magnitude of this difference must be evaluated relative to the variability of the term's frequency across the corpus and is therefore standardized by its standard deviation ($m_k$ . $s_i$). The final weight $w'_{ij}$ associated with the $i$th term in the $j$th novel is given in Equation 5, where Δ denotes the shrinkage parameter and $(v)_+$ is a function that returns v when v > 0 and zero otherwise.

$$\mathrm{w}'_{ij} = \mathrm{sign}(w_{ij}) \cdot \left(|w_{ij}| - \Delta\right) + \tag{5}$$

According to this formula, all term weights are reduced by the same value Δ. Moreover, a feature weight is set to zero when its value is smaller than Δ. Such features are considered to introduce noise rather than provide useful information for authorship attribution.

Increasing the value of Δ therefore reduces the number of features taken into account. In general, a given feature is useful for discriminating among only a few possible authors (or categories), rather than among all authors. Consequently, the $w'_{ij}$ values for a small number of authors may remain non-zero. However, when all corresponding weights $w'_{ij}$ for the *i*th term are set to zero, this feature is ignored for all subsequent attributions. This occurs when the overall mean frequency of the term is very similar across all authors, indicating that the term has little or no discriminative power.

**Table 6.** Ranked lists achieved by the NSC (200 and 500 MFWs) to Sara's novel (profile-based approach)

| | 200 MFW | | | 500 MFWs | | |
|---|---|---|---|---|---|---|
| Rank | Distance | % Change | Author | Distance | % Change | Author |
| 1 | 98.0 | | Nordbo | 292.6 | | Nordbo |
| 2 | 204.3 | 108.4% | Halfdan | 601.8 | 105.6% | Halfdan |
| 3 | 244.8 | 19.8% | Jensen | 638.4 | 6.1% | Jensen |
| 4 | 246.5 | 0.7% | Jessen | 670.3 | 5.0% | Sabroe |
| 5 | 259.1 | 5.1% | Sabroe | 672.8 | 0.4% | Jessen |

After applying this approach with 200 or 500 MFWs (Δ = 0.2), the resulting author rankings are reported in Table 6. Similar results are obtained when considering 100, 250, 300, or 800 MFWs. Moreover, varying the parameter Δ from 0.2 to 0.8 leads to the same outcome. These consistent results across different parameter settings provide strong evidence that Nordbo was involved in the writing of Sara Omar's novel.

An interesting aspect of the NSC method is its ability to reduce the number of features used to compute the similarity between the query document and the candidate authors. When considering 200 stylistic features and setting Δ = 0.2, 31 features (or 15.5%) are ignored during the similarity computation. Similarly, when Δ is increased to 0.5, 70 features are removed, representing 35% of the initial feature set. With 500 MFWs and Δ = 0.2, 87 features (or 17.4%) are discarded from the similarity calculation, and this reduction increases to 180 features (36%) when Δ is raised to 0.5. Clearly, within the set of MFWs, an appropriate feature-selection procedure can substantially reduce the computational burden without decreasing the effectiveness of the authorship attribution (Savoy, 2020).

Fourth, as a typical text classifier based on the machine-learning paradigm, the naïve Bayes model (Hastie *et al*., 2009; James *et al*., 2021) has been selected. This method was the first to be proposed for solving authorship attribution problems (Mosteller & Wallace, 1964), and it is commonly used as an effective baseline for evaluating machine-learning algorithms (James *et al*., 2021). Other studies have also proposed applying naïve Bayes to authorship attribution, including Juola and Vescovi (2011) and Savoy (2012), in which this method proved highly effective.

Given a set of possible authors (or hypotheses), denoted by $A_k$ for $k = 1, 2, \ldots, r$, the naïve Bayes model combines the prior probability that a given author wrote the query text, denoted by Prob[$A_k$], with the corresponding likelihood. The latter is defined as the probability of observing all terms $t_i$, for $i = 1, 2, \ldots, m$, given that the text was written by author $A_k$. This formulation assumes that the terms are conditionally independent given the author, an assumption that is generally unrealistic and gives the model its "naïve" designation. For a query novel $Q$, the naïve Bayes model selects as the most probable author the writer who maximizes Equation 6. In this equation, $t_i$ represents the $i$th term occurring in the doubtful text $Q$, while $n_Q$ denotes the size of this book.

$$Arg\ max_{A_k}\ \text{Prob}[A_k \mid Q] = \text{Prob}[A_k] \cdot \prod_{i=1}^{n_Q} \text{Prob}[t_i \mid A_k] \quad (6)$$

To estimate the prior probabilities, Prob[$A_k$], one can choose either a uniform distribution over all possible authors or a distribution based on the proportion of novels written by each author.

To determine the term probabilities, the corresponding author profile is taken into account. For each term $t_i$, this probability is estimated as the ratio of its frequency of occurrence in the profile of author $A_k$, denoted by $tf_{ik}$, to the size of this profile, $n_k$, as shown in Equation 7.

$$\text{Prob}[t_i \mid A_k] = {}^{\text{tf}_{ik}}\!/_{n_k} \quad (7)$$

This definition tends, however, to assign zero probability to terms that do not occur in the text. For such missing terms, the occurrence frequency (and hence the estimated probability) is zero, which can cause problems when computing the product of term probabilities. A smoothing approach must therefore be applied to address this issue. As with the other methods, we apply Lidstone's smoothing, which estimates

$$\text{Prob}[t_i|A_k] = (tf_{ik}+\lambda) / (n_k+\lambda\cdot|V|) \quad (8)$$

where $\lambda$ is a smoothing parameter (set to 0.1 in this study) and $|V|$ denotes the vocabulary size.

**Table 7.** Ranked lists achieved by the naïve Bayes model (200 and 500 MFWs) to Sara's novel (profile-based approach)

| | 200 MFW | | | 500 MFWs | | |
|---|---|---|---|---|---|---|
| Rank | Prob. | % Change | Author | Prob. | % Change | Author |
| 1 | 0.0139 | | Nordbo | 0.0087 | | Nordbo |
| 2 | 0.0134 | 3.28% | Halfdan | 0.0082 | 5.40% | Halfdan |
| 3 | 0.0133 | 0.96% | Korsgaard | 0.0081 | 1.07% | Korsgaard |
| 4 | 0.0132 | 0.79% | Gazan | 0.0080 | 0.71% | Sabroe |
| 5 | 0.0131 | 0.66% | Sabroe | 0.0080 | 0.54% | Gazan |

After applying this approach with 200 or 500 MFWs ($\lambda = 0.1$), the resulting author rankings are reported in Table 7. In these experiments, a uniform distribution was selected. Varying the number of stylistic features to 250, 300, or 800 MFWs produces the same results. Among all writers, Nordbo is consistently identified as the most likely author. The consistency of this finding across different feature-set sizes provides further evidence that Nordbo may have played a substantial role in the composition of Sara Omar's novel. As observed previously with the Delta model, the estimated probability decreases substantially from the first-ranked author to the second-ranked writer, suggesting that the first-ranked candidate, Nordbo, provides a substantially better fit to the novel than the other candidates.

## 6 Rolling Delta

In the previous experiments, Nordbo consistently emerged as the most likely author in the rankings generated by two authorship attribution methods. However, an important question remains: what was the extent of Nordbo's contribution? Was Nordbo the sole author of Sara's novel, or did he contribute as a coauthor? Under the first hypothesis, one would expect Nordbo's stylistic features to be present throughout the novel. Under the second hypothesis, the authorship attribution models might detect greater stylistic distances from Nordbo's profile in some sections of the book. Such outlying passages could indicate the possible presence of a second author.

To assess which of these two hypotheses is better supported by the data, the rolling Delta model was applied (Eder, 2016). Rather than analyzing the entire novel as a single document, the text was subdivided into chunks, each containing 5,000 tokens (words, numbers, and punctuation symbols). In a round-robin procedure, the writing style of each segment was compared with the profiles of the 15 novelists, excluding the Blædel–Nordbo co-authored novel.

More precisely, Sara's novel was divided into 20 non-overlapping chunks of 5,000 tokens each. The Delta model was then applied using either the 200 or 500 most frequent word-types (MFWs) as stylistic features. Consequently, 20 separate rankings were generated. To provide an overall view of the results, Figure 4 displays the estimated probability assigned to Nordbo for each segment. A dotted horizontal line indicates a probability of 0.5. The mean probability values obtained with the two MFW settings—0.748 for 200 MFWs and 0.629 for 500 MFWs—are also shown as dotted lines.

For the 200 MFW setting, the mean estimated probability assigned to Nordbo is higher than that obtained with 500 MFWs. For many segments, the estimated probability exceeds 0.8. A similar overall pattern is observed with 500 MFWs. However, several segments yield substantially lower probabilities, suggesting that their stylistic characteristics are less consistent with Nordbo's profile and may indicate the possible involvement of another author.

With the 200 MFW setting, the eighth segment shows a particularly low estimated probability for Nordbo (0.40). A similar pattern occurs for segments 14, 15, and 17, for which the estimated probabilities assigned to Nordbo are only 0.31, 0.28, and 0.43, respectively, although Nordbo remains the highest-ranked candidate in each case. These points are depicted with a small square in Figure 4. Thus, although Nordbo is consistently identified as the most likely candidate, the relatively low estimated probabilities for these segments do not provide strong support for attributing them to Nordbo.

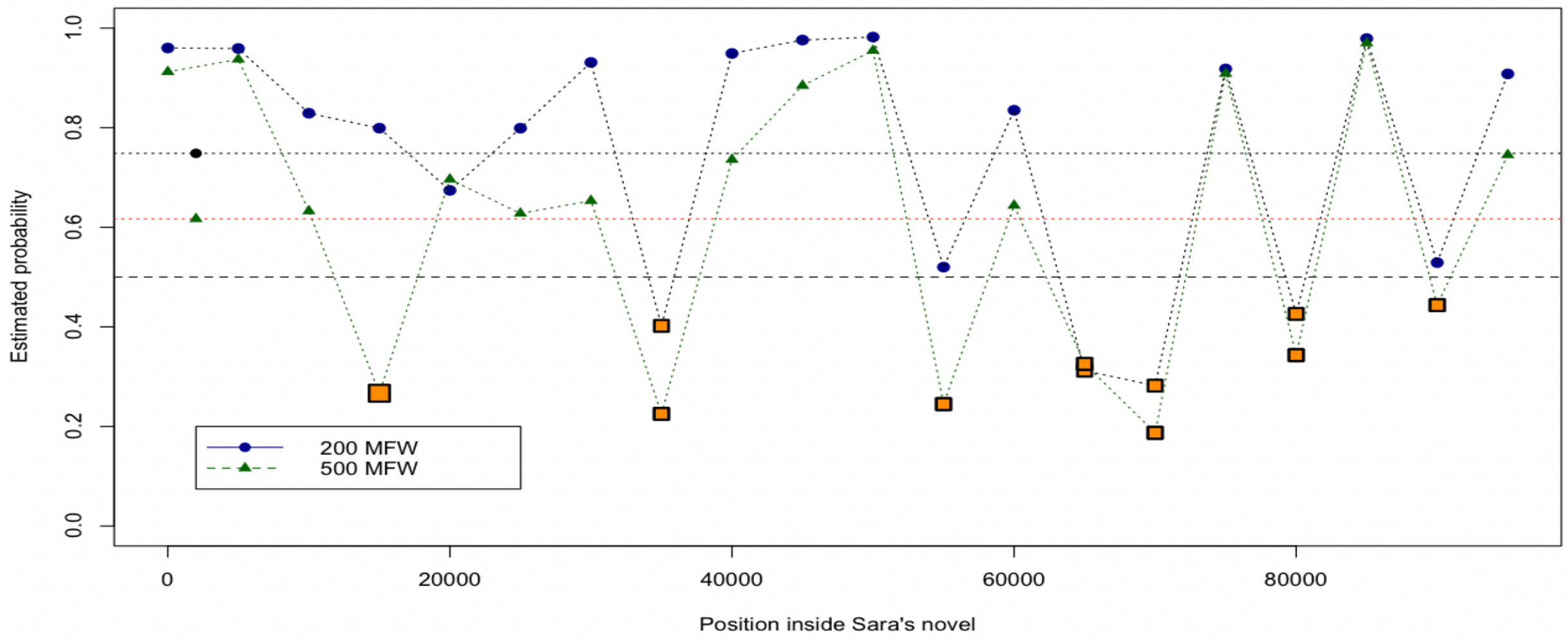

**Figure 4.** Rolling Delta with 200 and 500 MFWs (profile-based)

With the 500 MFW setting, Delta identifies Jensen as the most likely author for the fourth segment (probability: 0.51), followed by Nordbo (0.27). This case is indicated in large square in Figure 4. The eighth segment also yields a low estimated probability for Nordbo (0.23), although he remains the most likely candidate for this segment. As in the previous analysis, segments 14 and 15 yield very low estimated probabilities for Nordbo (0.32 and 0.19, respectively), although he remains the highest-ranked candidate. Similarly, Nordbo receives a relatively low estimated probability for the 12th segment (0.25). However, the success rate of the Delta method is not perfect, and its result should not be interpreted as DNA evidence.

To further examine these findings, the intertextual distance approach (Labbé, 2007) was adapted. First, instead of using an author profile, the shortest novel available for each author was selected. This choice was made to satisfy one of the constraints of Labbé's distance measure, namely that the difference in length between the two texts should not exceed a factor of eight. Since each segment of the doubtful novel contains 5,000 tokens, the lengths reported in Table A.1 indicate that this constraint is satisfied for most authors. The intertextual distance analysis consequently identifies Nordbo as the most similar author for the majority of Sara's segments.

However, for the problematic segments already identified by the rolling Delta analysis, the intertextual distance measure points to an author other than Nordbo. For example, for segments 12, 14, and 15, Jessen is identified as the most similar author.

Several observations can therefore be made. First, both authorship attribution methods identify the same segments as problematic. Second, these results are difficult to reconcile with the hypothesis that Nordbo was the sole contributor to the entire novel, since several segments exhibit substantially weaker stylistic similarity to his profile. At the same time, the overall results consistently identify Nordbo as the closest stylistic match for the majority of the novel, providing evidence of his involvement in its writing.

## 7 Some Linguistics Considerations

Based on the Danish corpus, our experiments with the authorship attribution models provide evidence that Nordbo was involved in the writing of Sara's novel. However, it is useful to complement this assessment with additional analyses examining potential linguistic or, at least, lexical similarities between Nordbo's books and the doubtful novel.

As a first approach, the most frequent words (MFWs) for each author can be examined. The results, however, are inconclusive, mainly because these MFW lists largely reflect the list presented in Section 2. For example, the most frequent words in Nordbo's novels are *og* (and, 3.4%), *i* (I, in, 2.6%), *det* (the, 2.5%), *han* (he, 1.9%), and *var* (was, 1.9%). Although the rankings differ slightly between writers, it is difficult to identify a meaningful explanation for these differences. In addition, author-specific MFW lists often include, among their higher-ranked entries, the names of characters appearing in their novels (e.g., Matthew for Nordbo, Elliot for Halfdan, and Marie for Gazan).

As a second approach, the *Z* scores computed for each word can identify terms used more frequently by a given writer than the mean frequency observed for all other authors. These values can be computed according to Formula 1, and Table 8 presents examples for three authors. Unlike the Delta model, however, these values are calculated using the entire vocabulary. The top entries in Table 8 reveal each author's preference for certain terms. Nevertheless, this information appears to reflect more the specific contexts of the different novels than clear stylistic features or a close linguistic relationship between Nordbo and Sara.

**Table 8**. Words overused by two authors (Nordbo & Sara Omar) and the coauthored novel

| Rank | Nordbo | Sara | Blædel-Nordbo |
|---|---|---|---|
| 1 | sneen (snow) | koranen (the Quran) | vers (verse) |
| 2 | dræbt (killed) | koran (Quran) | værkstedet (the workshop) |
| 3 | ravet (the sea) | khador (chador) | personer (people) |
| 4 | riflen (the riffle) | kvinder (women) | vandpolo (water polo) |
| 5 | isen (the ice) | koskéen (mosque) | citater (quotes) |
| 6 | bjørn (bear) | pige (girl) | præstegården (the vicarage) |
| 7 | båden (the boat) | hendes (her) | gylletank (slurry tank) |
| 8 | dræbte (killed) | vaske (wash) | forsvundne (missing) |
| 9 | riffel (rifle) | sagde (said) | golf (golf) |
| 10 | tæt (tight) | grandfætter (second cousin) | muslimer (muslins) |

As a more direct method for identifying close stylistic similarities, examining the contribution of individual words to the Delta distance may be useful. Table 9 presents the ten tokens with the most similar Z scores for Nordbo and Sara when considering 200 MFWs. The next two columns report their relative occurrence frequencies in Nordbo's and Sara's novels, respectively. As can be observed, the absolute differences between the two authors are relatively small. More precisely, the sum of these differences is 0.083%. To provide a better interpretation of this value, the last column of Table 9 reports the relative occurrence frequencies for Halfdan, a writer who exhibits some similarities to Sara, as shown in Table 2a. The sum of these differences is 8.56%, indicating a substantially greater discrepancy than that observed between Nordbo and Sara.

Of course, considering a larger set of terms makes it possible to complement the words presented in Table 9. Among these lexical items, the following have similar *Z* scores in both Nordbo's and Sara's texts: *hovedet* ("the head"), *fortsatte* ("continued"), *krop* ("body"), *råbte* ("shouted"), *hånd* ("hand"), *tanker* ("thoughts"), *øjne* ("eyes"), *fingre* ("fingers"), *vredt* ("angrily"), and *blod* ("blood"). These and other related terms can also be identified as having greater weights in the NSC method, indicating that they are highly discriminative lexical items. This group of words suggests a possible stylistic relationship, characterised in part by vocabulary associated with body parts and physical actions.

**Table 9**. The most similar tokens used to compute the Delta score

| Rank | Token | Nordbo | Sara | Halfdan |
|---|---|---|---|---|
| 1 | ind (in) | 0.44% | 0.44% | 0.29% |
| 2 | mens (while) | 0.20% | 0.20% | 0.10% |
| 3 | under (under) | 0.11% | 0.11% | 0.07% |
| 4 | så (so) | 1.37% | 1.35% | 0.94% |
| 5 | nikkede (noddded) | 0.15% | 0.16% | 0.05% |
| 6 | dem (them) | 0.32% | 0.33% | 0.29% |
| 7 | kiggede (looked) | 0.16% | 0.15% | 0.20% |
| 8 | sammen (together) | 0.19% | 0.19% | 0.09% |
| 9 | hvis (if) | 0.23% | 0.22% | 0.30% |
| 10 | se (to see) | 0.23% | 0.23% | 0.16% |

## 8 Conclusion

In investigating the authorship of the novel *Dødevaskeren* (*The Dead Washer*), published in 2017 under the name Sara Omar, the experiments presented in this study provide substantial evidence of Nordbo's involvement and a substantial contribution into the writing process. This finding is supported by four authorship attribution models that have been successfully evaluated in other contexts. Moreover, varying the parameter settings of these methods leads to consistent results.

When Nordbo is excluded from the set of possible authors, the authorship attribution models are unable to identify another plausible candidate. This supplementary experiment therefore provides additional support for the hypothesis of Nordbo's involvement. When examining possible close lexical relationships between Nordbo's profile and Sara's novel, an overuse of several terms related to the body can be observed, including *hand* (*hånd*), *eyes* (*øjne*), *fingers* (*fingre*), and *blood* (*blod*). Focusing on pronouns, the first-person personal pronoun *I* (*jeg*), as well as several forms of the third-person pronouns (e.g., *sig, sine, hans, dets,* and *deres*), occur with similar frequencies in Sara's novel and Nordbo's profile.

Moreover, the reference corpus provides relatively broad coverage of contemporary Danish novelists. For each selected writer, the corpus contains recent novels exceeding 10,000 words in length, providing sufficient textual material to derive reliable author profiles.

The main limitation of the present set of experiments is the absence of a novel by Sara Omar belonging to the same genre as Nordbo's books[5]. Thus, the possibility that Sara Omar is the author of the doubtful novel cannot be fully ruled out. However, in this case, the close correspondence between Nordbo's stylistic profile and extended passages (5,000 tokens) of the doubtful novel would require further explanation (see Figures 1 and 4). At the same time, Nordbo's stylistic contribution does not appear to extend uniformly throughout the entire novel, as some segments exhibit substantially weaker similarities to his stylistic profile. The rolling Delta and the intertextual distance confirm that for some segments Norddo' stylistic idiosyncrasies are not clearly present, suggesting the contribution of a second writer (particularly with chunk #12, #14, and #15 in Figure 4).

These findings should therefore be further examined using additional authorship attribution approaches and alternative sets of stylistic features, with different feature-set sizes and text representations, such as short sequences of letters or words. Such experiments could help determine whether the observed similarities with Nordbo are consistent across different methods and feature representations, and could provide further insight into the possible extent of his contribution to the novel.

Moreover, any automatic authorship attribution should ideally be complemented by a plain-English explanation that provides linguistic evidence supporting the proposed attribution. Such evidence could be based on the use of specific words or short sequences of words. However, a more informative approach would be to examine stylistic characteristics such as the distribution of parts of speech, sentence length, or recurrent sentence and clause constructions.

[5] To be precise, we do not have a novel written entirely by Sara Omar without external assistance or collaboration.

## Appendix

In Table A.1, a more complete information about the Danish corpus is reported. A tag corresponding to the author is associated to each book, together with an integer to discriminate between novels written by the same writer. Then novels are sorted in alphabetic order of their author. In the following columns, the title is given with the subgenre of the novel and its publication year. Finally, the length in tokens is displayed in the last column. The last two rows provide the information about the book coauthored (Blædel-Nordbo) and the doubtful novel (Sara Omar).

**Table A.1.** Complete information about our Danish corpus

| Author | Title | Subgenre | Year | Length (in tokens) |
|---|---|---|---|---|
| Blædel1 | *Bedemandens datter* | Suspense / Family | 2016 | 108,271 |
| Blædel2 | *Pigen under træet* | Crime | 2019 | 105,055 |
| Conradsen1 | *Centrifugen* | Crime | 2017 | 130,871 |
| Conradsen2 | *Familien Bruchfontaine* | Family saga | 2019 | 113,969 |
| Gazan1 | *Kvartsgeode* | Science / Suspense | 2020 | 16,958 |
| Gazan2 | *Svalens graf* | Science / Suspense | 2013 | 197,333 |
| Gerhardt1 | *Der bor Hollywoodstjerner på vejen* | Autofiction | 2014 | 42,593 |
| Gerhardt2 | *Transfervindue – fortællinger om de raskes fejl* | Autofiction | 2017 | 14,034 |
| Guldager1 | *En uskyldig familie* | Family | 2017 | 112,885 |
| Guldager2 | *Et rigtigt liv* | Family | 2019 | 81,597 |
| Halfdan1 | *Et nødvendigt offe* | Thriller | 2015 | 142,380 |
| Halfdan2 | *De uskyldige døde* | Crime | 2017 | 139,226 |
| Harvig1 | *Bedrag* | Crime / Psycho. | 2011 | 96,960 |
| Harvig2 | *Manden der vendte tilbage* | Crime / Psycho | 2017 | 89,610 |
| Høeg1 | *Tour de chambre* | Contemporary | 2020 | 27,176 |
| Høeg2 | *Nye rejsende* | Contemporary | 2017 | 13,603 |
| Jensen1 | *På kanten* | Historical | 2017 | 47,444 |
| Jensen2 | *Overleveren* | Historical | 2017 | **171,973** |
| Jessen1 | *Doktor Bagges anagrammer* | Contemporary | 2017 | 35,229 |
| Jessen2 | *En ny tid* | Historical | 2015 | 51,787 |
| Korsgaard1 | *Hvis der skulle komme et menneske forbi* | Contemporary | 2017 | 85,141 |
| Korsgaard2 | *En dag vil vi grine af det* | Contemporary | 2018 | 80,493 |
| Nordbo1 | *Pigen uden hud* | Crime / Thriller | 2017 | 110,748 |
| Nordbo2 | *Kold angst* | Crime / Thriller | 2018 | 112,535 |
| Nordbo3 | *Kvinden med dødsmasken* | Crime / Thriller | 2019 | 101,847 |
| Petersen1 | *Værelse til leje* | Crime | 2017 | 87,761 |
| Petersen2 | *Ønskebarnet* | Psycho. Suspense | 2009 | 95,913 |
| Sabroe1 | *32 centimeter* | Contemporary | 2018 | 114,540 |
| Sabroe2 | *Love Me Do* | Contemporary | 2017 | 107,886 |
| Tellerup1 | *Markløs* | Contemporary | 2017 | 47,788 |
| Tellerup2 | *Hedeselskabet* | Contemporary | 2020 | 51,291 |
| Blædel-Nordbo | *Opløst* | Crime / Thriller | 2021 | 110,110 |
| Sara Omar | *Dødevaskeren* | Family saga / Drama | 2017 | 100,300 |